\documentclass[letterpaper, 10 pt, conference]{ieeeconf}  

\IEEEoverridecommandlockouts                              

\usepackage{balance} 
\usepackage{amssymb,mathtools} 
\usepackage{gensymb} 
\usepackage{tabularx}
\usepackage{multirow} 
\usepackage{booktabs,tabulary,lipsum}

\usepackage{graphicx} 
\usepackage{caption}
\usepackage{subcaption}
\usepackage{tcolorbox}

\usepackage{amsmath}

\usepackage{hyperref}
\usepackage{cleveref}
\usepackage[ruled,vlined, norelsize]{algorithm2e}
\usepackage{cleveref}
\crefname{table}{Tab.}{Tabs.}
\crefname{figure}{Fig.}{Figs.}
\crefname{section}{Sec.}{Secs.}
\crefname{equation}{Eq.}{Eqs.}

\title{\LARGE \bf
Effects of Explanation Specificity on Passengers in Autonomous Driving
}

\author{Daniel Omeiza$^{1}$, Raunak Bhattacharyya$^{1}$, Nick Hawes$^{1}$, Marina Jirotka$^{2}$, Lars Kunze$^{1}$
\thanks{$^{1}$Oxford Robotics Institute,
        University of Oxford. Corresponding~Email: {\tt\footnotesize danielomeiza@robots.ox.ac.uk}}%
\thanks{$^{2}$Dept. of Computer Science, University of Oxford.}%
}

\SetKwInput{KwInput}{Input}
\SetKwInput{KwOutput}{Output}

\newcommand{\red}[1]{\textcolor{red}{#1}}
\newcommand{\blue}[1]{\textcolor{blue}{#1}}
\newcommand{\green}[1]{\textcolor{teal}{#1}}

\begin{document}

\maketitle
\thispagestyle{empty}
\pagestyle{empty}

\newcommand\copyrighttext{
  \footnotesize ©2022 IEEE. Personal use of this material is permitted. Permission from IEEE must be obtained for all other uses, in any current or future media, including reprinting/republishing this material for advertising or promotional purposes, creating new collective works, for resale or redistribution to servers or lists, or
reuse of any copyrighted component of this work in other works.}

\newcommand\copyrightnotice{
\begin{tikzpicture}[remember picture,overlay]
\node[anchor=south,yshift=10pt] at (current page.south) {\fbox{\parbox{\dimexpr\textwidth-\fboxsep-\fboxrule\relax}{\copyrighttext}}};
\end{tikzpicture}%
}

\begin{abstract}
The nature of explanations provided by an explainable AI algorithm has been a topic of interest in the explainable AI and human-computer interaction community. In this paper, we investigate the effects of natural language explanations' specificity on passengers in autonomous driving. We extended an existing data-driven tree-based explainer algorithm by adding a rule-based option for explanation generation. We generated auditory natural language explanations with different levels of specificity (\textit{abstract} and \textit{specific}) and tested these explanations in a within-subject user study (N=39) using an immersive physical driving simulation setup. Our results showed that both abstract and specific explanations had similar positive effects on passengers' perceived safety and the feeling of anxiety. However, the specific explanations influenced the desire of passengers to takeover driving control from the autonomous vehicle (AV), while the abstract explanations did not. We conclude that natural language auditory explanations are useful for passengers in autonomous driving, and their specificity levels could influence how much in-vehicle participants would wish to be in control of the driving activity.

\end{abstract}
\section{Introduction}
The automotive industry is witnessing an increasing level of development in the past decades, from manufacturing manually operated vehicles to manufacturing vehicles with a high level of automation.
As highly automated vehicles make high-stake decisions that can significantly affect end-users, the vehicles should explain or justify their decisions to meet set transparency guidelines or regulations.

Associating natural language explanations with an AV's driving decisions is one promising approach for better vehicle transparency~\cite{omeiza2021explanations}. This transparency, obtained through intelligible explanations, can help to reassure passengers of safety and also assist them in effectively calibrating their trust in an AV~\cite{khastgir2018calibrating}. The specificity level of explanations is, however, important in achieving the aforementioned benefits. For example, while vehicle operators, developers, and incident investigators might desire very specific and detailed explanations from an AV for auditing and debugging purposes, it's not clear what impact such level of specificity would have on passengers. Would very specific explanations that are capable of exposing AV errors be beneficial to passengers?

Further, as passengers are expected to be able to engage in other activities during an autonomous ride, the visual mode of communicating awareness to passengers might be futile in conditions where a human is required to intervene. Hence, other feedback mechanisms such as auditory communication~\cite{kunze2019conveying} are needed.
\begin{figure}
     \centering
         \includegraphics[width=6.5cm]{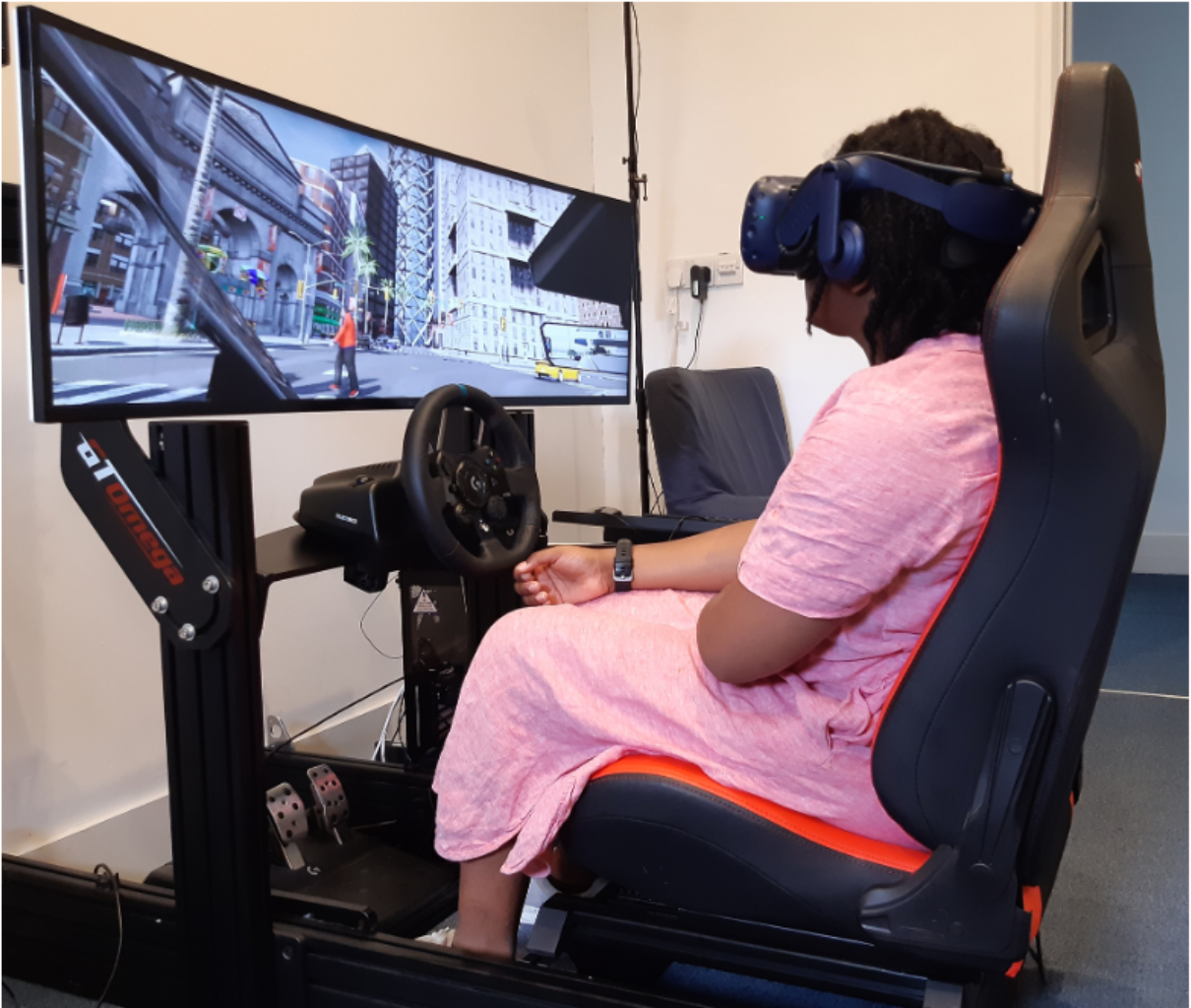}
         \label{fig:participant}
\caption{\footnotesize Driving simulation setup for the study. The setup included a VR headset, steering wheel, brake and acceleration pedals, screen, and arcade seat. The screen shows a pedestrian crossing at a crosswalk.}~\label{fig:setup}
\end{figure}

In this study, we use an immersive driving simulator, an automated auditory explainer, and a virtual reality headset to investigate the effects of explanation specificity on passengers in highly automated vehicles. The effects of interest are perceived safety, the feeling of anxiety, and the feeling to takeover control from an AV. While there are related works on external human-machine interfaces~\cite{liu2021importance}, we focus on auditory explanations provided to in-vehicle participants.

We use the term \textit{abstract} to mean the provision of vague auditory explanations that conceal some details about a driving situation. The term \textit{specific} is used to mean the provision of very specific explanations with more details about a situation. Our contributions are:
\begin{enumerate}
    \item a use case of explanation specificity in the autonomous driving context;
    \item an enhanced interpretable technique for generating auditory natural language explanations for AV navigation actions;
    \item findings on whether high AV transparency, though critical to other stakeholders, is helpful to AV passengers.
\end{enumerate}

\section{Related Work}
\label{sec:rel_work}
Explanations have been found useful in enhancing user experience~\cite{schneider2021explain}, trust~\cite{koo2015did, ha2020effects}, and improved situational awareness~\cite{omeiza2021not,liu2021importance} in automated driving. Recent works have explored human factors in the application of explainable AI in autonomous driving. For instance, in~\cite{omeiza2021towards, omeiza2022spoken}, a socio-technical approach to explainability was proposed. An interpretable representation and algorithms for explanations based on a combination of actions, observations, and road rules were designed. 

In relation to explanation depths, the ideology that explanations with higher abstractions and/or correctness are better has been discussed in~\cite{buijsman2022defining, guidotti2018survey}.
Ramon et al.~\cite{ramon2021understanding} also argued that explanation specificity depends on the application context, and in particular, low-level specificity is preferred for people with a more deliberative cognitive style.

In this paper, the term \textit{explanation specificity}
is used to refer to two specificity levels of explanations, \textit{abstract} (low transparency) and \textit{specific} (high transparency). Explanations can be used to convey different information in autonomous driving, e.g., vehicle uncertainties and intentions, and communicated through different modalities. For example,
Kunze et al.~\cite{kunze2019automation} conveyed visual uncertainties with multiple levels to operators using heartbeat animation. This information helped operators calibrate their trust in automation and increased their situation awareness. Similarly, Kunze et al.~\cite{kunze2019conveying} used peripheral awareness display to communicate uncertainties with the aim of alleviating the workload on operators simultaneously observing the instrument cluster and focusing on the road. This uncertainty communication style decreased workload and improved takeover performance. In addition, the effects of augmented reality visualisation methods on trust, situation awareness, and cognitive load have been investigated in previous studies using semantic segmentation~\cite{colley2021effects}, scene detection and prediction~\cite{colley2022effects}, and  pedestrian detection and prediction~\cite{colley2020effect}. These deep vision-based techniques applied to automated driving videos and rendered in augmented reality mode were a way of calling the attention of operators to risky traffic agents in order to enhance safety. While under-explored, auditory means of communicating explanations are important to calling in-vehicle participants' attention to critical situations in autonomous driving. We thus used an auditory communication style in this study to convey explanations to passengers.
Some existing works around human-machine interaction~\cite{liu2021importance} have leveraged theoretical models (e.g., mental and situational models~\cite{endsley2000situation}) to study explanations. We based our work on behavioural cues and subjective feedback from subjects while drawing connections to such existing works.

\section{Passenger Study}
\label{sec:userstudy}
In this section, we describe the participants' demography, experiment apparatus setup, experiment design, and the procedure of the experiment. The necessary approval to conduct the study was obtained from the University of Oxford Research Ethics Committee.
\subsection{Participants}
We conducted a power analysis to estimate the number of subjects required for the study. Afterwards, calls for participants were placed on various online platforms, such as the callforparticipants platform, university mailing groups, university Slack channels, the research group website, and social media to recruit subjects. Upon screening, the final sample consisted of $N = 39$ participants (28 male, 11 female) ranging in age from 18 to 59 years.
The participants comprised students, university employees, and members of the callforparticipants platform. Although prior driving experiences were not required, 28 (71.79 \%) of the participants were licensed drivers.
Only 2 of the 39 participants (5.13\%) had experience with autonomous vehicles, however, in a research context. 6 (15.38\%) of the participants had used a virtual reality headset for a driving game or driving experiment in the past.

{\fontsize{8pt}{10pt}
\selectfont
\begin{table*}[]
\centering
\caption{\footnotesize Description of a subset of the events (5 out of 9) and corresponding explanations provided during the study. Observations and causal explanations are announced to passengers. AV's action (text in red), other agent's class \& action (text in blue), and the agent's location (text in green) are determined by the explainer algorithm described in Algorithm~\ref{alg:tb}.}

\begin{tabular}{@{}p{2.5cm} p{4.5cm} p{3.5cm} p{3.5cm}@{}}
\toprule
\textbf{Event} &
  \textbf{Description} &
  \textbf{Observation} &
  \textbf{Causal Explanation} \\ \midrule
FollowLeadingVehicle &
  AV follows a leading actor. At some point, the leading actor slows down and finally stops. The AV has to react accordingly to avoid a collision. &
  \blue{vehicle ahead} on \green{my lane}. &
  \red{Stopping} because \blue{cyclist stopped} on \green{my lane}. \\
VehicleTurning &
  AV takes a right or a left turn from an intersection where an actor suddenly drives into the way of the AV,  AV stops accordingly. After some time, the actor clears the road, AV continues driving. &
  \blue{motorbike crossing} \green{my lane}. &
  \red{Stopping} because \blue{motorbike is crossing} \green{my lane}. \\
LaneChangeObstacle &
  AV follows a leading actor, and at some point, the leading actor decelerates. The AV reacts accordingly by indicating and then changing lanes. &
  \blue{vehicle ahead} on \green{my lane}. &
  \red{Changing lane to the {[}right/left{]}} because \blue{vehicle stopped} on \green{my lane}. \\
StopSignalNoActor &
  No actor ahead of the AV at a signalised intersection with a red traffic signal. AV decelerates and stops. &
  \blue{red traffic light} ahead on \green{my lane}. &
  \red{Stopping} because \blue{traffic light is red} on \green{my lane}. \\
MovSignalNoActor &
  No actor ahead of the AV. AV starts moving from a stop state at a signalised junction or intersection. &
  None &
  \red{Moving} because \blue{traffic light is green} on \green{my lane}. \\
\bottomrule
\end{tabular}
\label{tab:scenarios}
\end{table*}
}
\subsection{Apparatus}
\subsubsection{Hardware}
The hardware setup is shown in \Cref{fig:setup}.
We conducted the experiment in a driving simulator that comprised a GTR arcade seat, Logitech G29 steering wheel with force feedback, turn signal paddles, brake and accelerator pedals, and an ultra-wide LG curved screen to display the experiment. A state-of-the-art virtual reality (VR) headset (with an immersive $360^{\circ}$ FoV and an eye tracker) was also used  to provide an immersive experience and high visual fidelity.
\begin{figure}
\centering
  \includegraphics[height=4.7cm]{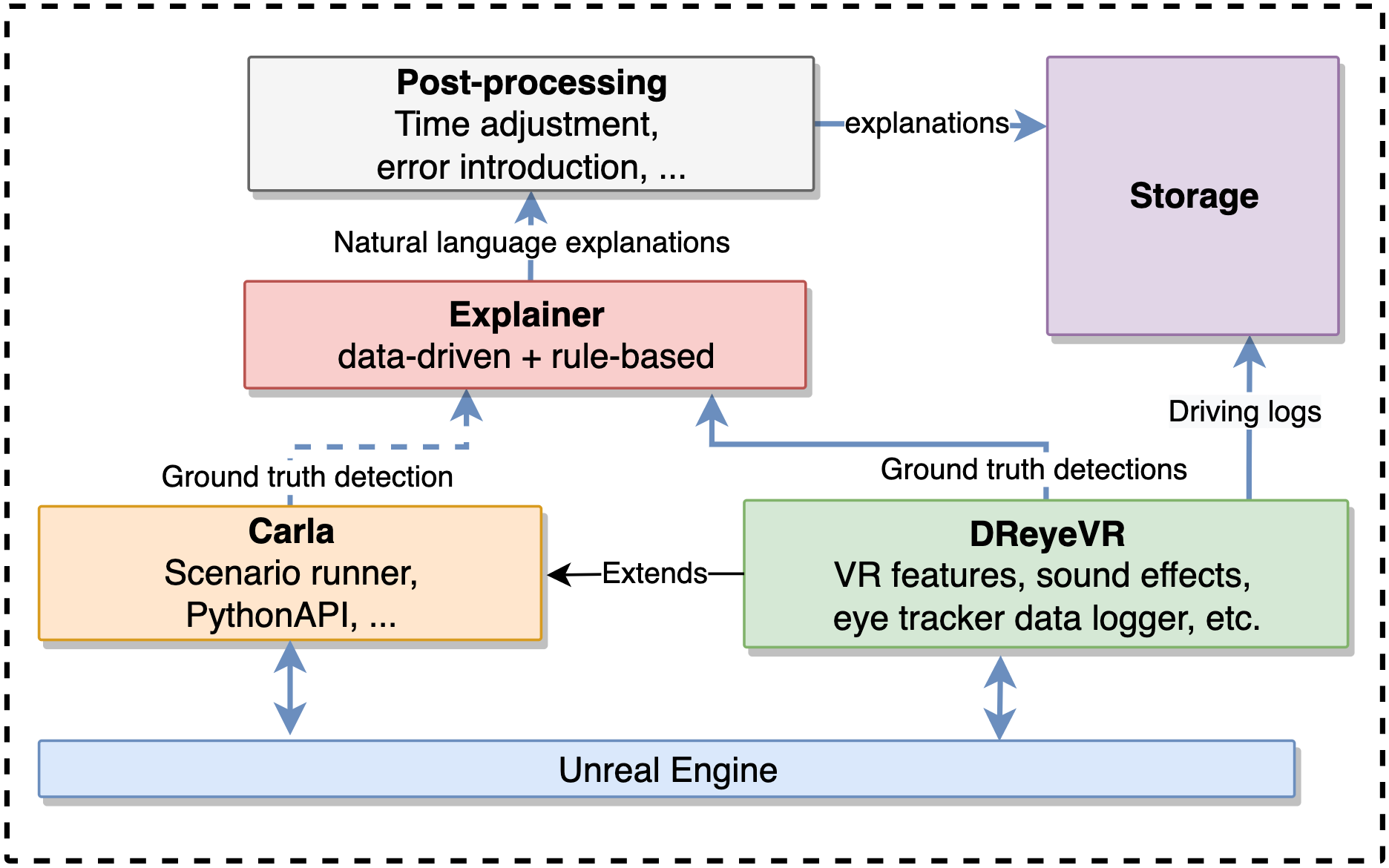}
  
  \caption{ \footnotesize High-level architecture of our simulation software. DReyeVR uses Unreal engine and extends  Carla simulator, which also builds on Unreal engine. DReyeVR extends Carla by adding VR functionalities, vehicular and ambience sounds, eye tracker data logging, and additional sensors, among others. Our explainer model, which is both rule-based and data-driven, receives ground truth data from  Carla or DReyeVR and generates explanations for predicted actions. The post-processing script allowed us to modify the generated explanations as we desire.}~\label{fig:arch}
\end{figure}

\subsubsection{Driving Software}
Software architecture is illustrated in \Cref{fig:arch}.
We adapted the DReyeVR~\cite{silvera2022dreyevr}, an open-source VR-based
driving simulation platform for behavioural and interaction research involving human drivers. DReyeVR was built atop Carla~\cite{dosovitskiy2017carla}, an open-source driving simulator for autonomous driving, and Unreal Engine 4. DReyeVR provides a very realistic experience with naturalistic visuals (e.g., in-vehicle mirrors) and auditory (e.g. vehicular and ambient sounds) interfaces allowing for an ecologically valid setup. It also provides an experimental monitoring and logging system to record and replay scenarios, as well as a sign-based navigation system.

\subsubsection{Explainer Software}
As shown in \Cref{fig:arch}, we developed an explainer system (based on previous work in~\cite{omeiza2022spoken}) that uses a tree-based model fitted on an AV driving dataset that we have collected and annotated (with a multilevel annotation scheme) in a previous project. While the original algorithm in~\cite{omeiza2022spoken} is mainly data-driven, we incorporated a rule-based technique that acts as a fallback when the data-driven method fails or makes an incorrect ego action prediction. While the data-driven method uses a trained tree-based model to predict and generate explanations from the detections from Carla, the rule-based approach uses Carla's ground truth data and follows pre-defined rules to determine which agent(s) to include in the explanation. With the data-driven approach, we are able to know when a prediction is incorrect by comparisons with ground truth observations from Carla's simulation data. We used this improved explainer system (data-driven and rule-based) to generate preliminary explanations for our created scenarios.  
While Wintersberger et al.~\cite{wintersberger2020explainable} suggested the types of traffic elements to be included in visual explanations based on a study on user preferences, our proposed explainer picks up traffic elements that the driving model deemed important (cf. ~\cite{anjomshoae2021context}) for its driving decisions (see Algorithm~\ref{alg:tb}).

We performed post-processing operations on the generated explanations. Post-processing operations included fine-tuning some of the explanations and modifying the explanations' timestamps to make them come at the right time.

\begin{algorithm}
\LinesNumbered
\DontPrintSemicolon
\KwInput{ tree model $\mathcal{M}$ for ego's action prediction, input vector $X$ describing ego's environment}
\KwOutput{intelligible auditory explanation}
    Select a representative tree  $m$ from tree model $\mathcal{M}$.\\
    Predict action $y \in \mathcal{Y}$ given $X$.\\
    Compare prediction with Carla ground truth.\\
    If prediction matches Carla ground truth, goto step 5 otherwise use Carla ground truth information and predefined rules to generate explanation following the template in \Cref{tab:scenarios} and end process.\\
    Trace the decision path for the prediction $y$ in tree $m$.\\
    Compute the importance score of the attributes in each node along the decision path.\\
    Select attributes with importance scores $ \ge $ some threshold $k$.\\
    Merge the conditions/inequalities in the selected attributes. \\
    Translate merged attributes to natural language following the template in \Cref{tab:scenarios}.\\
   
\caption{Intelligible Explanation Generation}
\label{alg:tb}
\end{algorithm}

\subsection{Experiment Design}
Before the start of the trials, participants manually drove a vehicle in VR mode for about two minutes in Carla Town03. Thirty vehicles and ten pedestrians were spawned in this town. The aim of the drive was to familiarise participants with the simulation environment.

A within-subject design was then implemented with one independent variable: \textit{specificity}, and three dependent variables: \textit{perceived safety, feeling of anxiety}, and \textit{takeover feeling}. The first specificity level (\textit{abstract} comprised \textit{vague explanations that can conceal all the AV's perception errors}. The second specificity level (\textit{specific}) comprised \textit{more specific and detailed explanations indicating high transparency}. A within-subject design was chosen to avoid any potential co-founding factor of between-individual differences in a between-subject design. We didn't have a control scenario where explanations were not provided because the goal of the study was to investigate the impact of explanation specificity and not the presence of explanations. Previous studies have already shown that explanations, including placebo explanations that convey no helpful information, provide positive effects on people~\cite{eiband2019impact}. Hence, we focused on how the specificity of these explanations influences passengers.
\subsubsection{Independent Variables}
We created two driving scenarios, one in which abstract explanations were provided and the other with specific explanations. The driving  scenarios were carefully designed to include different driving conditions that are obtainable in the real world (See \Cref{tab:scenarios}). 

\paragraph{Abstract Scenario} a route from Carla Town10HD,  which was about 4 minutes in length (330 secs), was created. Driving conditions were a combination of the events in \Cref{tab:scenarios}. The rules governing explanations for this scenario were:
(i) all traffic lights are referred to as `traffic signs' without specifying the state (e.g., red, green, amber, off) of the traffic light (ii) pedestrians are referred to as `road users' (iii) all non-human moving actors are referred to as `vehicle'. This includes cycles, motorbikes, cars, etc.
An example explanation is `stopping because of the traffic sign on my lane'. This obfuscates the type or colour of the traffic sign.

\paragraph{Specific Scenario} A scenario in Carla Town10HD, which was about 4 minutes in length (256 seconds), was created. Driving conditions in this scenario were also a combination of the events in \Cref{tab:scenarios}. The explanations generated in this scenario were fine-grained, and detailed and could expose any perception system errors in the AV. We introduced 5\% error into the perception system of the AV as an attempt to model a realistic AV perception system. This error value was estimated following the dynamic traffic agent classification model and confusion matrix provided in~\cite{bin2021double}. We were only interested in the confusion matrices (and not the models). The confusion matrices helped us to systematically introduce the 5\% perception system errors to be reflected in the specific explanations. This amounted to one erroneous explanation out of the 22 explanations provided in this scenario. An example of an erroneous explanation is: `van ahead on my lane'. Here, a car was misclassified as a van. Note that this error was insignificant to the AV's navigation actions.

We counterbalanced the routes across scenarios. That is, the AV's route was different in each scenario. This design decision was made to reduce carry-over effects on the participants. With this setup, the scenarios were still comparable as they were all within the same town, and the routes shared similar features. Each scenario also had a balanced combination of the events listed in Table~\ref{tab:scenarios}. In both scenarios, the AV maintained a speed below $30mph$, the recommended speed limit in urban areas in the UK. The AV also respected all road rules and avoided collisions in both scenarios.

\subsubsection{Dependent Variables}
The \textit{Autonomous Vehicle Acceptance Model Questionnaire (AVAM)}~\cite{hewitt2019assessing} was adopted to assess \textit{perceived safety} and the \textit{feeling of anxiety} dependent variables. AVAM is a user acceptance model for autonomous vehicles, adapted from existing user acceptance models for generic technologies. It comprises a 26-item questionnaire on a 7-point Likert scale, developed after a survey conducted to evaluate six different autonomy scenarios. We selected Items 19---21 to assess the feeling of anxiety factor and Items 24---26 to assess the perceived safety factor. 

Similar to~\cite{schneider2021explain}, we introduced a new item to assess participants' feeling to takeover navigation control from the AV during the ride (\textit{takeover feeling}). Specifically, participants were asked to rate the statement `During the ride, I had the feeling to take over control from the vehicle' on a 7-point Likert scale. Actual navigation takeover by participants was not permitted because we wanted to be able to control the entire experiment and have all participants experience the same scenarios. Moreover, we were dealing with L4 automation. Though participants were not expected to drive or take over control, they might have nursed the thought to do so. This is what the \textit{takeover feeling} dependent variable measured.

We also added a free response question directly related to explanations. Participants were asked the following question: `What is your thought on the explanations provided by the vehicle, e.g., made you less/more anxious, safe, feeling to take over control?'. We refer to the resulting questionnaire as the APT Questionnaire (i.e., A-Feeling of Anxiety, P-Perceived Safety, T-Takeover Feeling).

\begin{figure}
\centering
\includegraphics[width=\columnwidth]{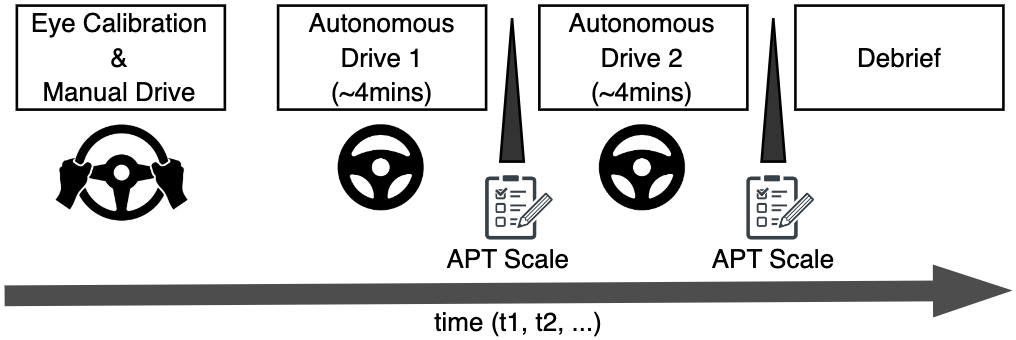}
  \caption{ \footnotesize Study procedure. Eye calibration is done with the VR headset; participants drive for two minutes, participants experience each of the ~4 mins scenarios in counterbalanced order and complete the A-Feeling of Anxiety, P-Perceived Safety, T-Takeover Feeling questionnaire (APT Scale) in between each scenario. Participants are debriefed.}~\label{fig:flow}
\end{figure}

\subsection{Procedure}
The procedure of the experiment is illustrated in \Cref{fig:flow}. After all preliminary form completions and briefings, we introduced the physical driving simulator and explained the next steps, which involved a pre-experiment manual driving session (in VR mode) which lasted for 2 minutes. The participants were informed that the purpose of the pre-experiment exercise was to help them get familiar with the simulation environment. This exercise also helped us to exclude those with motion sickness from the actual experiment.

When the manual driving exercise was completed, we took the VR headset off the participant and explained the aim and the procedure of the main experiment. ``you would experience two autonomous rides in different vehicles, [...] and after each ride, you would complete a short survey. The vehicle drives along a predefined path for about 4 minutes and provides explanations for its planned driving decisions, and announces relevant objects in its environment [...]. The vehicle tells you its next direction at a junction or an intersection using its right or left red light indicators on its dashboard accordingly. [...] Simply click any of these buttons if the decision or the explanation of the vehicle makes you feel confused, anxious or unsafe. Note that you cannot take over driving control from the vehicle during the drives''. The researcher then puts the VR headset back on the participants and launched the two driving scenarios (one after the other) in a complete counterbalanced order.

\section{Quantitative Results}
\begin{figure}
\centering
  \includegraphics[height=5.3cm]{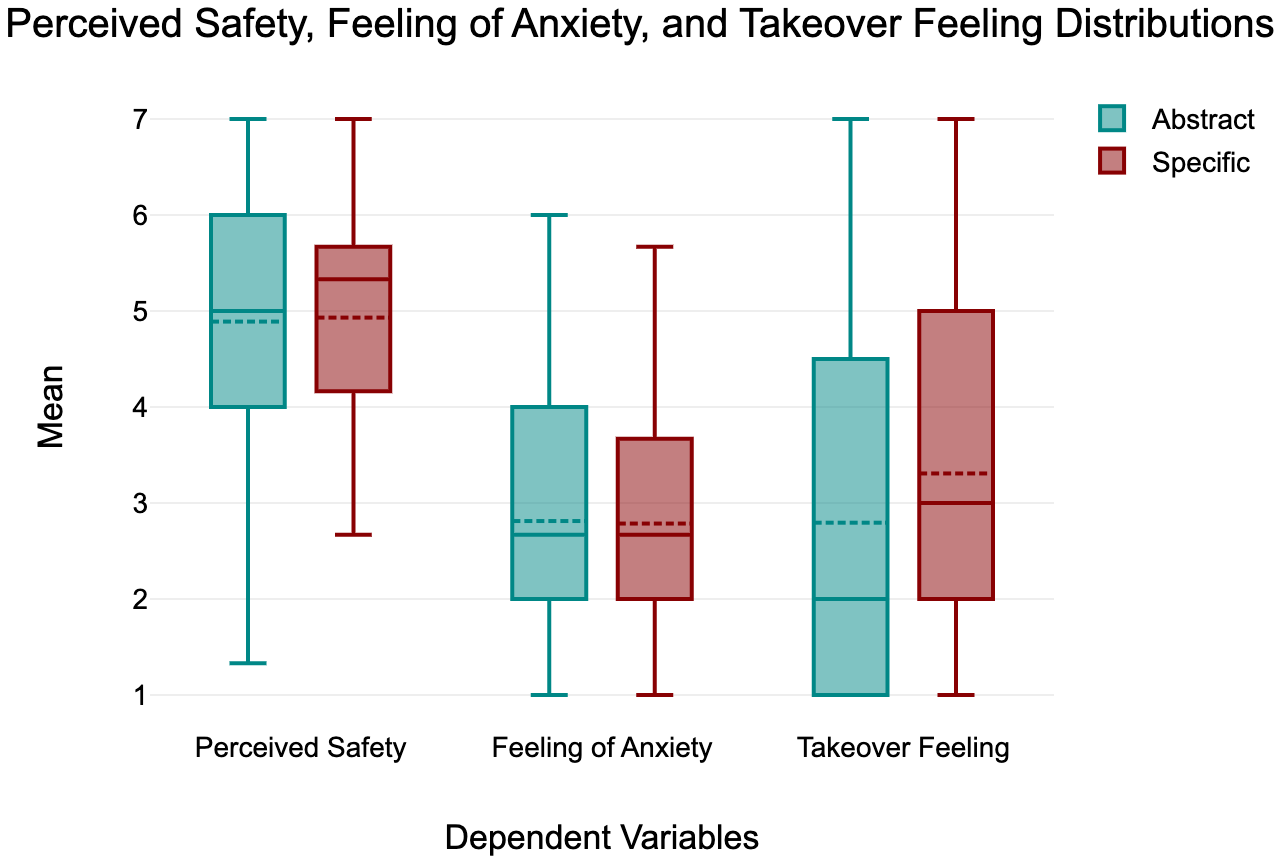}
  \caption{\footnotesize The plot shows the summary statistics of the dependent variables. \textit{Specific} explanations yielded a higher perceived safety and takeover feeling.}~\label{fig:factors}
\end{figure}

We aim to investigate the effect of explanation specificity on passengers' perceived safety, the feeling of anxiety, and takeover feeling. We analysed the data from the two APT questionnaires administered. A latent variable (perceived safety) was formed from the means of the responses from AVAM Items 24---26 to assess participants' perceived safety during the study. Another latent variable (anxiety feeling) was formed from the means of AVAM Items 19---21. 
We calculated the Cronbach Alpha ($\alpha$) for the independent variables to see if they had adequate internal consistency.
Takeover feeling was also assessed using the 7-point Likert scale question introduced into the APT questionnaire.
Results with p-value less than 0.05 ($p < .05$) are reported as significant. Bonferroni corrections were done in all statistical tests to reduce the chance of Type 1 errors.
Kolmogorov-Smirnov, Shapiro-Wilk, and Anderson-Darling tests indicated a violation of normality for the perceived safety, feeling of anxiety,  and takeover feeling variables. Therefore, the Friedman test was performed for these dependent variables (see \Cref{fig:factors}).

\begin{table*}[]
\centering
\caption{\footnotesize Descriptive statistics from APT questionnaire analysis. $H(2)$ denotes Chi-square value}.
\begin{tabular}{lcccccccccccl}\toprule
& \multicolumn{3}{c}{\parbox{3cm}{\textbf{Perceived Safety} \\Cronbach $\alpha: 0.87$, \\${H(2)} = 0.03, \boldsymbol{p=.872}$}} & \multicolumn{3}{c}{\parbox{3cm}{\textbf{Feeling of Anxiety} \\Cronbach $\alpha: 0.86$\\ ${H(2)} = 0.641, \boldsymbol{p=.423}$}} & \multicolumn{3}{c}{\parbox{3cm}{\textbf{Takeover Feeling} \\ ${H(2)} = 4.33, \boldsymbol{p=.037}$}}
\\\cmidrule(lr){2-4}\cmidrule(lr){5-7} \cmidrule(lr){8-10}
& Mean  & SD & Mean Rank & Mean  & SD  & Mean Rank & Mean & SD  & Mean Rank &  \\\midrule
Abstract   & 4.89 & 1.35 & 2.15 & \textbf{2.81} & 1.34 & 1.72 & 2.79 & 1.91 & 1.68 \\
Specific & \textbf{4.93} & 1.13 & \textbf{2.22} & 2.79 & 
1.2 & \textbf{1.81} & \textbf{3.31} & 1.79 & \textbf{2.10} \\
\bottomrule
\end{tabular}
\label{tab:results}
\end{table*}

\subsubsection{Perceived Safety}
\textit{Specific scenario} had a higher mean rank of 2.22 compared to the \textit{abstract scenario} with a mean rank of 2.15. However, no significant statistical difference was observed in perceived safety across the  \textit{abstract} and \textit{specific scenario} cases when the Friedman test was performed (see~\Cref{tab:results}). While \textit{specific} explanations yielded a higher perception of safety in our experiment, this relative difference is statistically insignificant. 

\subsubsection{Feeling of Anxiety} 
\textit{Specific scenario} had a higher mean rank of 1.81 compared to the \textit{abstract scenario} with a mean rank of 1.72. Similar to the perceived safety result, no significant statistical difference was observed in the feeling of anxiety across the \textit{abstract} and \textit{specific} scenarios when the Friedman test was performed (see~\Cref{tab:results}). Hence, explanations specificity, are as well inconsequential to the feeling of anxiety in the context of the study.

\subsubsection{Takeover Feeling}
For the takeover feeling dependent variable, \textit{specific scenario} had a higher mean rank of 2.10 compared to the \textit{abstract scenario} with a mean rank of 1.68. The Friedman test indicated a significant statistical difference between the \textit{abstract scenario} and the 
\textit{specific scenario}, $H(2) = 4.23,  p=.037$ (see \Cref{tab:results}). Furthermore, our statistical analysis showed no statistically significant difference in the takeover feeling variable between those who possessed and those who did not possess a driving licence ($p > 0.05$). Hence, specific explanations could evoke more takeover thoughts in passengers than abstract explanations in an AV.

\begin{figure}
\centering
  \includegraphics[width=\columnwidth]{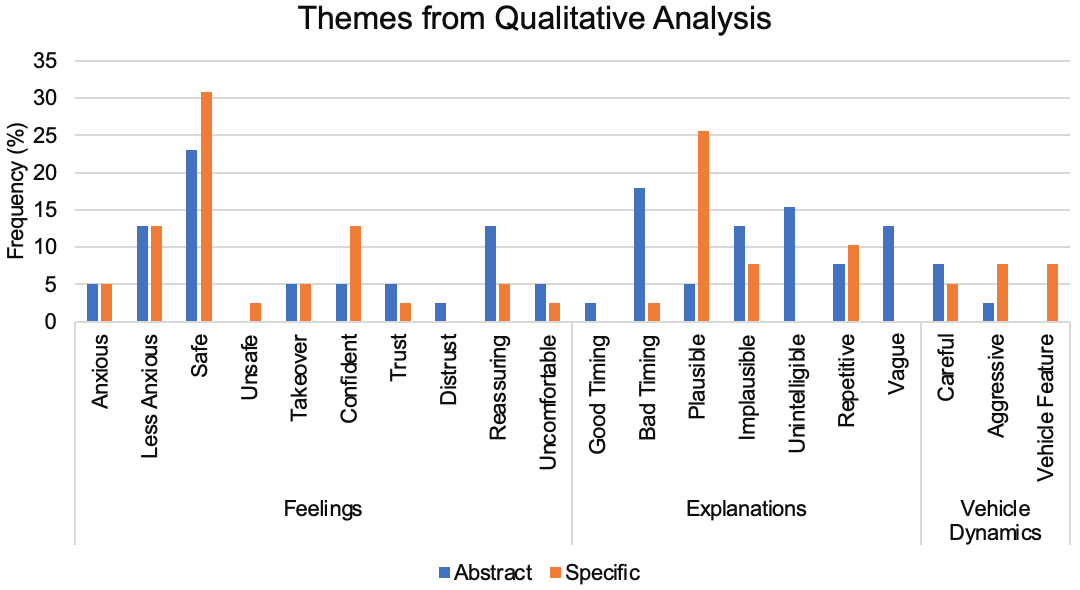}
  \caption{\footnotesize Themes derived from the thematic analysis of the qualitative data from participants. Frequency is expressed in percentage of the total number of responses in a scenario.}~\label{fig:feedback}
\end{figure}

\section{Qualitative Results: Themes and Reflections}
\label{sec:reflection}
We obtained qualitative data from the APT questionnaire administered after each scenario run. Participants were asked to describe their feelings with respect to the explanations that they received during the ride. \Cref{fig:feedback} describes the themes obtained from the inductive thematic analysis of the comments. Themes were broadly categorised based on the participants' feelings, their assessment of the explanations, and the vehicle dynamics.
\subsection{Feelings}
Both driving experiences (\textit{abstract} and \textit{specific}) produced positive effects on passengers' safety. Passengers felt safer mostly through the reassurance that the explanations provided. While the abstract explanations were a bit confusing to the passengers, they didn't create a significant negative impact on the passengers' perceived safety. See sample quotes:
\textit{`It was initially confusing due to the strange terminology used by the explanations. However, because the use of the explanations was consistent, it did inspire some confidence that the car was safe and knew what it was doing.'}---CAND25 (Abstract).

\textit{`[...] Safer and with more correct directions and decisions. Cyclist and motorcyclist wear no helmets.'}---CAND38 (Specific).

Comments regarding the feeling of anxiety seem to have an equal number of appearances in both the \textit{abstract} and \textit{specific} scenarios. Explanations in both cases made people less anxious.
\textit{`they probably contributed to make me feel less anxious. [...]'}---CAND31 (Abstract).
\textit{`Explanations were reassuring and made me feel less anxious.'}--CAND22 (Specific).

While many participants felt safe in both driving cases, a few nursed the thought to be in charge of the driving activity at some points, e.g., \textit{`I am okay with the vehicle driving because they don't make mistake. I don't feel unsafe but sometimes I feel like being in control. The explanations were simple.'}---CAND17 (Abstract).

Some participants preferred their own driving style to that of the AV, and for this reason, they felt like being in control at certain points, e.g., \textit{`Some of the car's decisions and corresponding explanations did not align with what I would have done in the situation and therefore made me feel like I would like to take over control.'}---CAND35 (Specific).

\subsection{Explanations}
Participants did notice the vagueness of the explanations in the \textit{abstract} scenario. Some thought it was good, while others thought it was confusing and made them uncomfortable.
\textit{`Its explanations were not specific enough since they only referred to traffic signs instead of the colour of the lights. This made me doubt the vehicle's actions a bit, even if they were correct.'}---CAND15 (Abstract).
\textit{`[...] Also I would be more comfortable if the explanation 'traffic sign' was 'traffic light is red/green' when referring to a traffic light.'}---CAND23 (Abstract).

There were more comments on the plausibility of the explanations in the \textit{specific} scenario compared to those in the \textit{abstract} scenario. \textit{`It explained the situation and its actions well, although sometimes it would perform an action and not provide an explanation (e.g. stopping briefly in front of a stop sign without voicing the action or situation). I still trust the vehicle's explanations since they were accurate descriptions of the situation.'}---CAND15 (Specific).
A couple of participants thought that the explanations in
the \textit{abstract} scenario were either too early or late. For
example, \textit{`The explanations should have arrived a bit earlier, like a few meters before the vehicle actually stops so that I will know that it is planning to stop. [...]'}---CAND23 (Abstract).

\subsubsection{Vehicle Dynamics}
Some comments were made about the vehicle's driving style and its interior. There was a comment relating to aggressive manoeuvre in the \textit{abstract} scenario: `\textit{Seemed like oncoming vehicles were going to collide with me. It seems to sometime drive on pavements when negotiating corners.}'---CAND29 (Abstract). 
The rotating steering wheel of the vehicle made some of the participants uncomfortable: `\textit{The steering wheel moving abruptly startled me sometimes.}'---CAND1 (Specific).

\section{Discussion}
Our results corroborate prior studies by showing that intelligible explanations create positive experiences for users in autonomous driving~\cite{omeiza2021towards, ha2020effects, schneider2021explain, m2021calibrating}.
While \textit{specific} explanations might provide details that are likely to expose perception errors, evidence from this study shows that these errors, when they are not consequential, have no significant effect on passengers. Passengers would feel safe as far as the AV makes the right decisions. In fact, \textit{specific} explanations tend to create a higher perception of safety (using the mean rank metric). This is against the thought that \textit{abstract explanations}, with their ability to abstract details, would hide possible errors from the passengers, providing a `higher' sense of safety. Moreover, placebo explanations have been shown to have positive effects as real explanations on people~\cite{eiband2019impact}.

A link between perceived safety and anxious feelings has been assumed in the literature~\cite{davidson2016mediating,quansah2022perceived}. Hence, since participants' perceived safety was highest in the \textit{specific} scenario, we expected a lower feeling of anxiety in the \textit{specific} scenario as well. Our results matched this expectation. Dillen et al.~\cite{dillen2020keep} observed that in-vehicle features, such as the rotating steering wheel, could influence the feeling of anxiety in in-vehicle participants. This was reinforced in our study by the comment from CAND1. We note that anxiety is hard to objectively capture, so the results from this experiment are only based on the participants' perceptions, and thus the term `feeling of anxiety'.

Although passengers were not meant to takeover control from the vehicle in this study, we expected that they would conceive the idea to do so when they repeatedly received vague  explanations that were not clear or too specific explanations that exposed subtle errors inherent in the AV. We found that takeover feeling was higher in the \textit{specific} scenario. This might be because the participants were able to better understand the reasoning process of the AV through the details that the explanations provided. This understanding allowed the participants to predict and judge the AV's actions, leading to the thought to takeover control where the actions of the AV were irreconcilable with the participants' driving preference. We observe from the qualitative results that the thought to takeover might not necessarily be triggered by errors but could be the participants' desire to be in the driving loop or their strong preference for their own driving style. This aligns with the argument  in~\cite{terken2020toward} that shared control rather than human-out-of-the-loop automated driving is required.

In general, explanations are helpful for passengers in autonomous driving. The level of explanations specificity might not have a significant effect on passengers' perceived safety and feeling of anxiety but might influence their thought to takeover control from an AV.

\section{Conclusion}
We conducted a within-subject lab study ($N=39$) using an immersive driving simulator to investigate the effects of explanation specificity on passengers' perceived safety, the feeling of anxiety, and takeover feeling.
Our results showed that both \textit{abstract} and \textit{specific} auditory natural language explanations are helpful for improving passengers' perceived safety and reducing the feeling of anxiety with no particular specificity level significantly better than the other. However, the specificity of the explanations influenced the passengers' thought to takeover control from the AV. In particular, more participants nursed the thought to takeover control from the AV at certain points when they received specific explanations. In future work, we will investigate the effect of varying degrees of AV perception system errors as an additional dimension to the independent variables explored in this study.

\section*{ACKNOWLEDGMENTS}
This work was supported by the EPSRC RAILS project (grant reference: EP/W011344/1) and the EPSRC project RoboTIPS (grant reference: EP/S005099/1).

\footnotesize{
\bibliographystyle{ieeetr}
\bibliography{bibliography}
}


\end{document}